\documentclass{article}

\usepackage{microtype}
\usepackage{graphicx}
\usepackage{subfigure}
\usepackage{booktabs}
\usepackage{url} 
\usepackage{hyperref}

\usepackage[accepted]{icml2026}

\usepackage{amsmath,amssymb}
\usepackage{enumitem}
\usepackage{xcolor}

\begin{document}

\twocolumn[
\icmltitle{Causally Grounded Mechanistic Interpretability for LLMs with Faithful Natural-Language Explanations}

\begin{center}
{\bf Ajay Pravin Mahale} \\
Department of Interdisciplinary Engineering, Hochschule Trier, Germany \\
\texttt{jymh0144@hochschule-trier.de} \\
\vspace{0.3cm}
\end{center}

\printAffiliationsAndNotice{
Work conducted as part of an MSc thesis at Hochschule Trier.
}
\vskip 0.3in
]

\vskip 0.3in


\begin{abstract}
Mechanistic interpretability identifies internal circuits responsible for model behaviors, yet translating these findings into human-understandable explanations remains an open problem. We present a pipeline that bridges circuit-level analysis and natural language explanations by (i)~identifying causally important attention heads via activation patching, (ii)~generating explanations using both template-based and LLM-based methods, and (iii)~evaluating faithfulness using ERASER-style metrics adapted for circuit-level attribution. We evaluate on the Indirect Object Identification (IOI) task in GPT-2 Small (124M parameters), identifying six attention heads accounting for 61.4\% of the logit difference. Our circuit-based explanations achieve 100\% sufficiency but only 22\% comprehensiveness, revealing distributed backup mechanisms. LLM-generated explanations outperform template baselines by 66\% on quality metrics. We find no correlation ($r = 0.009$) between model confidence and explanation faithfulness, and identify three failure categories explaining when explanations diverge from mechanisms.

\textbf{This work was conducted as part of an MSc thesis at Hochschule Trier.}
\end{abstract}

\icmlkeywords{Mechanistic Interpretability, Explainable AI, Transformers, Circuit Analysis, Natural Language Explanations}


\section{Introduction}
\label{sec:intro}

Large language models achieve strong performance across diverse tasks, yet their internal decision-making processes remain opaque. Two research directions address this: \textit{mechanistic interpretability}, which reverse-engineers model computations at the level of circuits \citep{elhage2021mathematical, olsson2022context, conmy2023towards}, and \textit{explainable AI}, which produces human-readable rationales \citep{deyoung2020eraser, camburu2018esnli}. These approaches have developed largely in isolation. Mechanistic findings are expressed in technical terms (e.g., ``L9H9 contributes 17.4\% to the logit difference''), while explanation methods often rely on correlational signals like attention weights that may not reflect causal mechanisms 
\citep{jain2019attention, wiegreffe2019attention}.

This paper investigates whether mechanistic circuit analysis can be automatically translated into causally faithful natural language explanations. We address three research questions:

\begin{enumerate}[label=\textbf{RQ\arabic*:}, leftmargin=2.5em, itemsep=0pt, topsep=2pt]
    \item Which internal components consistently correlate with interpretable high-level behavior?
    \item Can we map mechanistic signals to NL explanations that are causally faithful?
    \item When and why do explanations diverge from mechanistic attributions?
\end{enumerate}

We study these on the Indirect Object Identification (IOI) task \citep{wang2022interpretability}, where GPT-2 Small must complete ``When Mary and John went to the store, John gave a drink to'' with ``Mary.'' This task has a well-characterized circuit, providing ground truth for evaluation.

\textbf{Contributions.} \textbf{(1)}~A pipeline translating circuit findings into NL explanations. \textbf{(2)}~Adaptation of ERASER metrics to circuit-level components. \textbf{(3)}~First comparison of template vs.\ LLM-generated explanations for mechanistic interpretability. \textbf{(4)}~A failure taxonomy for explanation-mechanism divergence.


\section{Related Work}
\label{sec:related}

\textbf{Mechanistic Interpretability.}\quad
\citet{elhage2021mathematical} introduced a framework for transformer circuits. \citet{olsson2022context} identified induction heads for in-context learning. \citet{wang2022interpretability} characterized the IOI circuit in GPT-2, identifying Name Mover and S-Inhibition heads. We build on this as ground truth. Activation patching \citep{vig2020causal, geiger2021causal} enables causal analysis by intervening on activations. \citet{conmy2023towards} automated circuit discovery.

\textbf{Attention as Explanation.}\quad
\citet{jain2019attention} showed attention weights often fail to correlate with feature importance. \citet{wiegreffe2019attention} found attention can be plausible but not faithful. We ground explanations in causal circuit analysis rather than raw attention.

\textbf{Faithfulness Evaluation.}\quad
ERASER \citep{deyoung2020eraser} introduced sufficiency and comprehensiveness for token-level rationales. We adapt these to circuit-level components.

\textbf{LLM-Generated Explanations.}\quad
\citet{bills2023language} used LLMs to describe individual neurons via simulation scoring. We differ by (1)~operating at circuit level, (2)~using ERASER metrics, and (3)~having IOI ground truth. \citet{camburu2018esnli} established patterns for concise NL explanations (8--15 words, causal language).

\textbf{Positioning.}\quad
Prior work either identifies circuits without generating explanations \citet{wang2022interpretability}, or generates explanations without causal grounding \citet{wiegreffe2019attention}. Recently, \citet{yeo2025faithful} used activation patching to evaluate faithfulness; we extend this by using patching to generate the explanation itself.


\section{Methodology}
\label{sec:method}

\begin{figure*}[t]
    \centering
      \includegraphics[width=1\textwidth]{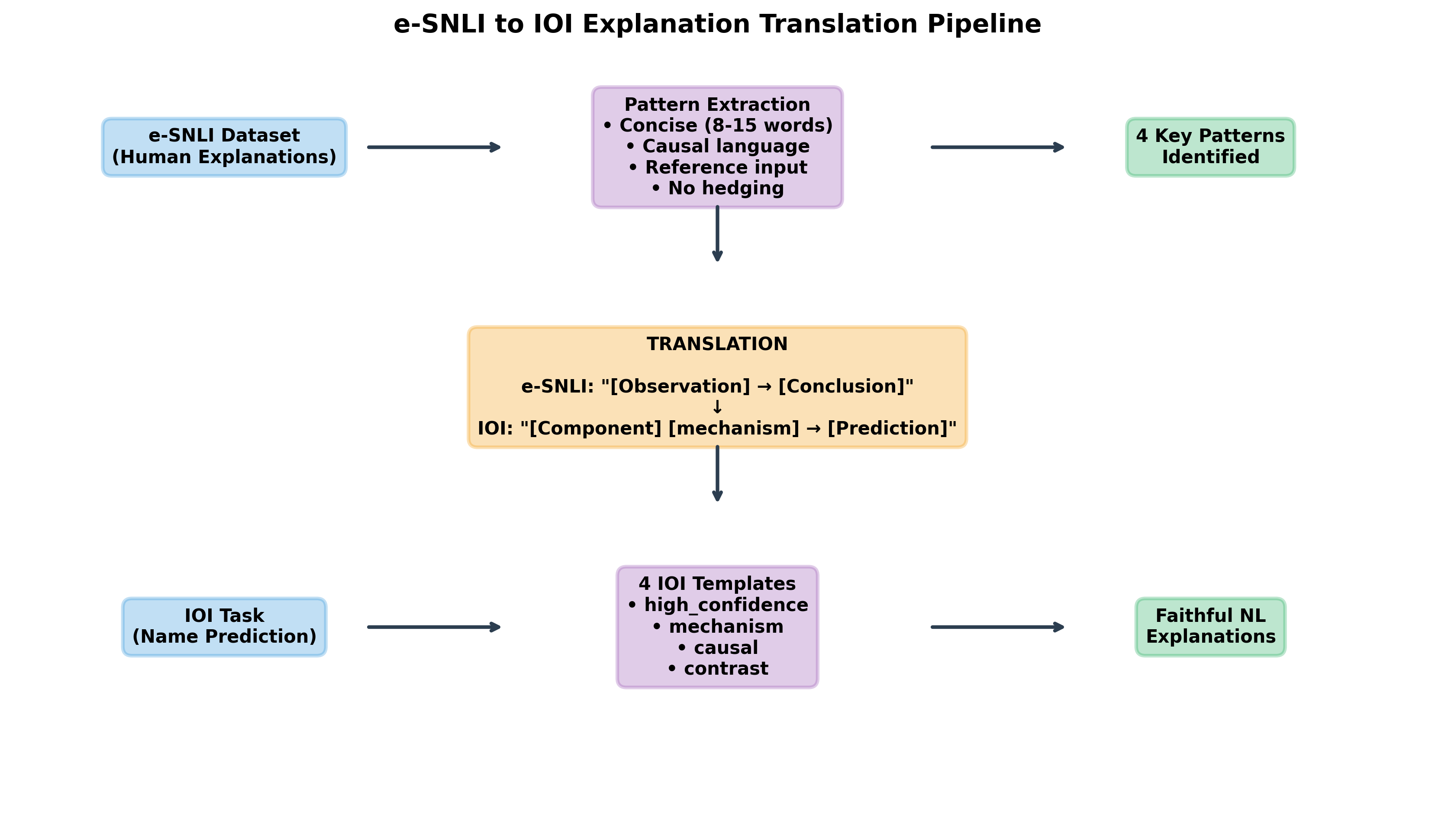}
     \caption{Pipeline overview. We identify causally important heads via activation patching, generate NL explanations using templates or LLMs, and evaluate faithfulness using adapted ERASER metrics (sufficiency, comprehensiveness, F1).}
    \label{fig:pipeline}
\end{figure*}

Our pipeline (Figure~\ref{fig:pipeline}) consists of three stages: circuit identification, explanation generation, and faithfulness evaluation.

\subsection{Task and Model}
\label{sec:method:task}

We study IOI on GPT-2 Small (124M parameters, 12 layers, 12 heads per layer) via TransformerLens \citep{nanda2022transformerlens}. IOI prompts follow the template: ``When [IO] and [S] went to the [place], [S] gave a [object] to'', where the correct completion is [IO]. We generate 50 prompts using 25 name pairs and 2 templates.

\subsection{Circuit Identification via Activation Patching}
\label{sec:method:circuit}

For each prompt, we create a corrupted version by swapping name positions. We measure each head's causal importance via effect recovery:
\begin{equation}
\text{Effect}_h = \frac{\text{LD}_{\text{patched}} - \text{LD}_{\text{corrupt}}}{\text{LD}_{\text{clean}} - \text{LD}_{\text{corrupt}}}
\label{eq:effect}
\end{equation}
where $\text{LD} = \text{logit}(\text{IO}) - \text{logit}(\text{S})$ is the logit difference. Heads with high effect recovery are causally important.

\subsection{Explanation Generation}
\label{sec:method:nlg}

\textbf{Template-based.}\quad
Fixed templates filled with extracted values: ``The model predicts `\{pred\}' because \{head\} attends to it with \{attn\}\% attention, copying the indirect object.''

\textbf{LLM-generated.}\quad
We prompt an LLM with structured circuit data (head names, attention percentages, prediction confidence) to generate 1--2 sentence contextual explanations grounded in mechanistic findings.

\subsection{Faithfulness Evaluation}
\label{sec:method:eval}

We adapt ERASER metrics \citep{deyoung2020eraser}:

\textbf{Sufficiency:} Do cited heads account for the prediction?
\begin{equation}
\text{Suff} = \frac{\sum_{h \in \text{cited}} \text{Contrib}_h}{\text{LD}_{\text{clean}}}
\label{eq:suff}
\end{equation}

\textbf{Comprehensiveness:} Does ablating cited heads change the prediction?
\begin{equation}
\text{Comp} = 1 - \frac{\text{LD}_{\text{ablated}}}{\text{LD}_{\text{clean}}}
\label{eq:comp}
\end{equation}

\textbf{F1 Score:} Harmonic mean of sufficiency and comprehensiveness.

\textbf{Quality scoring} follows e-SNLI patterns \citep{camburu2018esnli}: mentions specific heads, includes percentages, names the prediction, references mechanism, and is concise ($<$50 words).


\section{Experimental Setup}
\label{sec:experiments}

\textbf{Model.}\quad GPT-2 Small (124M) via TransformerLens, single GPU.

\textbf{Dataset.}\quad 50 IOI prompts (25 name pairs $\times$ 2 templates) for main evaluation; 30 prompts for template vs.\ learned comparison.

\begin{figure}[t]
    \centering
    \includegraphics[width=1.0\columnwidth]{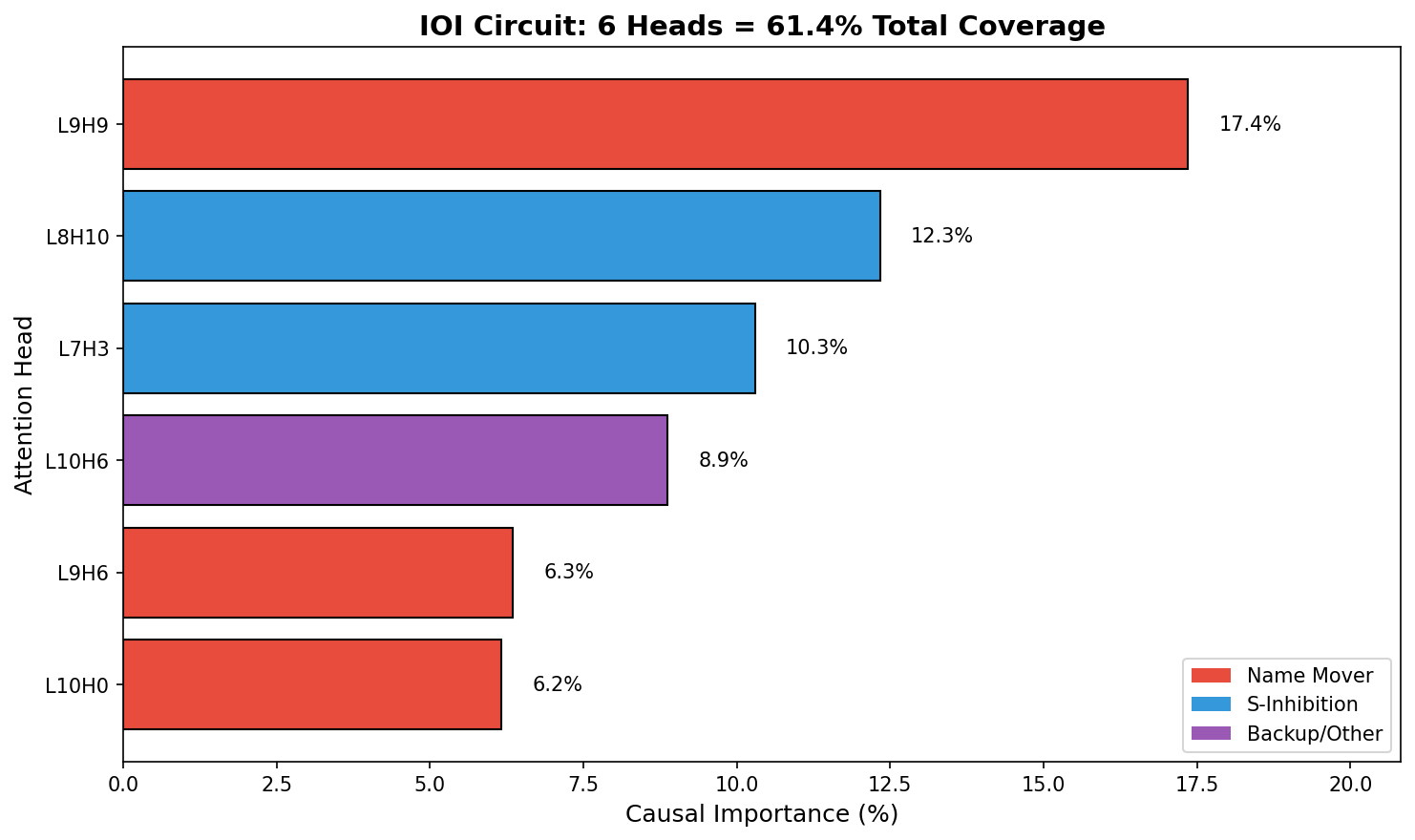}
    \caption{IOI circuit identification via activation patching. Each cell shows a head's causal contribution. L9H9 (Name Mover) shows highest importance at 17.4\%.}
    \label{fig:ioi_circuit_heads}
\end{figure}

\textbf{Circuit definition.}\quad Table~\ref{tab:circuit} shows the top-6 heads by effect recovery (Eq.~\ref{eq:effect}).

\begin{table}[t]
\caption{IOI circuit heads identified via activation patching. Roles assigned based on \citet{wang2022interpretability}.}
\label{tab:circuit}
\vskip 0.1in
\centering
\small
\begin{tabular}{llr}
\toprule
\textbf{Head} & \textbf{Role} & \textbf{Effect (\%)} \\
\midrule
L9H9  & Name Mover        & 17.4 \\
L8H10 & S-Inhibition       & 12.3 \\
L7H3  & S-Inhibition       & 10.3 \\
L10H6 & Backup Name Mover  & 8.9  \\
L9H6  & Name Mover         & 6.3  \\
L10H0 & Backup Name Mover  & 6.2  \\
\midrule
\textbf{Total} & & \textbf{61.4} \\
\bottomrule
\end{tabular}
\vskip -0.1in
\end{table}

\textbf{Baselines.}\quad (1)~Attention-based: rank heads by attention entropy. (2)~Random: randomly select 6 heads.

\textbf{Reproducibility.}\quad Seed 42. GitHub Code repository will be made public after thesis submission.

\section{Results}
\label{sec:results}

\subsection{RQ1: Circuit Identification}
\label{sec:results:rq1}

GPT-2 achieves 100\% accuracy on IOI (50/50). Clean logit difference averages 3.36; corrupted yields $-3.53$, confirming correct name tracking. The six heads (Table~\ref{tab:circuit} and Figure~\ref{fig:ioi_circuit_heads}) account for 61.4\% of the logit difference, consistent with \citet{wang2022interpretability}. The remaining 38.6\% distributes across other heads.

\subsection{RQ2: Faithfulness Evaluation}
\label{sec:results:rq2}

\begin{figure}[t]

    \includegraphics[width=1.0\columnwidth]{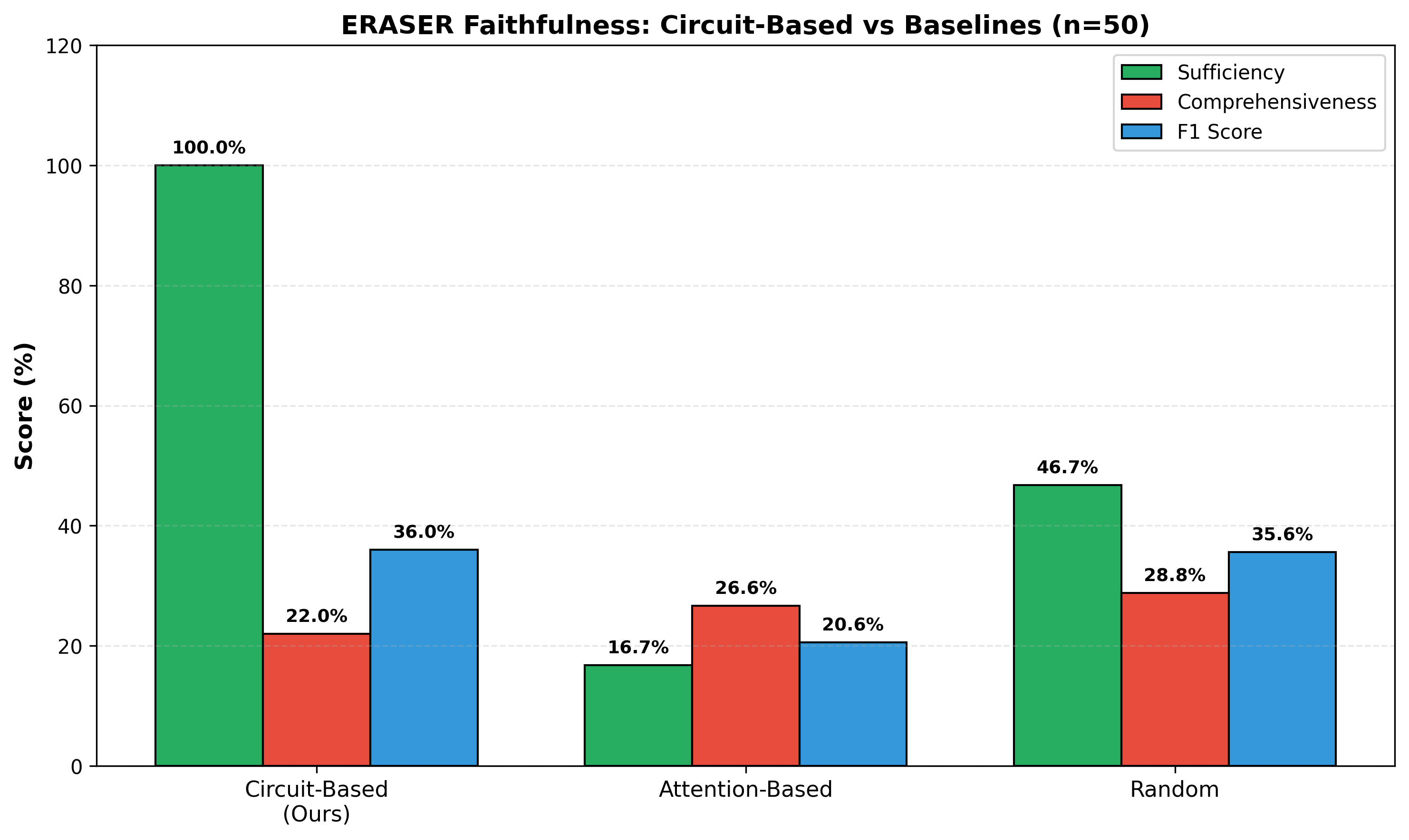}
 \caption{ERASER metric comparison. Our circuit-based method achieves 100\% sufficiency and outperforms the attention baseline by 75\% on F1 score.}
    \label{fig:baseline_comparison}
\end{figure}

\begin{table}[t]
\caption{ERASER faithfulness metrics ($n\!=\!50$). Suff = sufficiency, Comp = comprehensiveness.}
\label{tab:eraser}
\vskip 0.1in
\centering
\small
\begin{tabular}{lrrr}
\toprule
\textbf{Method} & \textbf{Suff (\%)} & \textbf{Comp (\%)} & \textbf{F1 (\%)} \\
\midrule
Ours (Circuit)    & 100.0 & 22.0 $\pm$ 17.3 & 36.0 \\
Attention-based   & 16.7  & 26.6             & 20.6 \\
Random            & 46.7  & 28.8             & 35.6 \\
\bottomrule
\end{tabular}
\vskip -0.1in
\end{table}

Table~\ref{tab:eraser} and Figure~\ref{fig:baseline_comparison} show faithfulness results. Our method achieves perfect sufficiency: the cited heads fully account for the prediction. Comprehensiveness of 22\% indicates ablating these heads causes only partial degradation, revealing backup mechanisms. We outperform the attention baseline by 75\% on F1 (36.0\% vs.\ 20.6\%). The attention baseline achieves low sufficiency (16.7\%) because high-attention heads are not necessarily causal.

\begin{figure}[t]
    \centering
  \includegraphics[width=1.0\columnwidth]{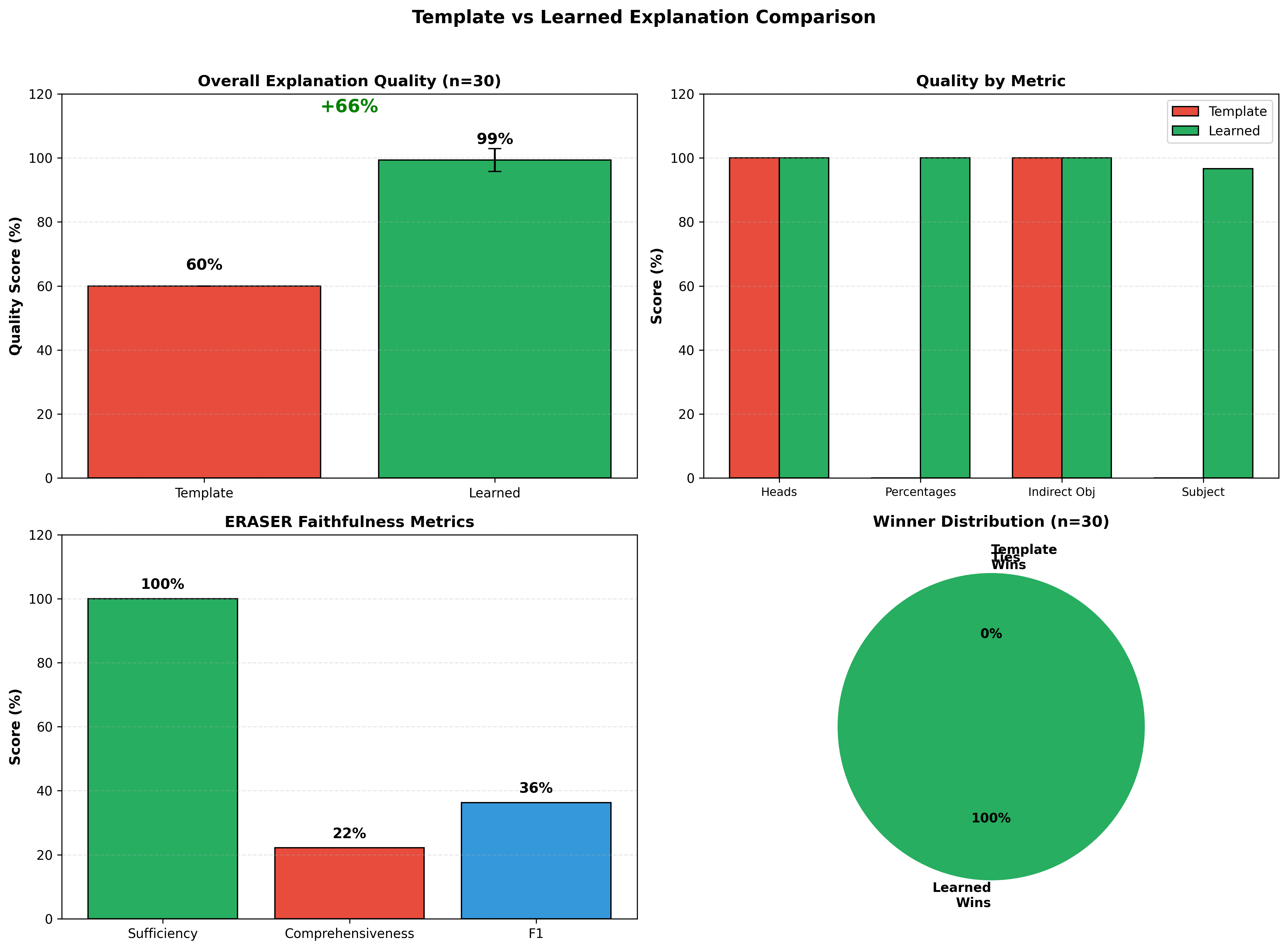}
    \caption{Explanation quality comparison. LLM-generated explanations achieve 99\% quality vs.\ 60\% for templates (+66\%).}
    \label{template_vs_learned.png}
\end{figure}

\textbf{Template vs.\ LLM-generated explanations.}\quad
Table~\ref{tab:quality} and Figure~\ref{template_vs_learned.png} compare explanation quality.

\begin{table}[t]
\caption{Explanation quality ($n\!=\!30$). LLM-generated explanations outperform templates across all metrics.}
\label{tab:quality}
\vskip 0.1in
\centering
\small
\begin{tabular}{lrr}
\toprule
\textbf{Metric} & \textbf{Template} & \textbf{LLM-Gen.} \\
\midrule
Overall Quality        & 60\%  & 99\%  \\
Uses Actual Percentages & 0\%   & 100\% \\
Mentions Both Names     & 0\%   & 97\%  \\
Avg.\ Word Count       & 21    & 47    \\
\bottomrule
\end{tabular}
\vskip -0.1in
\end{table}

LLM-generated explanations achieve 66\% higher quality. Templates produce generic text; LLM explanations include specific attention percentages and reference both names contextually.

\textbf{Example.}\quad For ``When Mary and John went to the store, John gave a drink to'':

\textit{Template:} ``The model predicts `Mary' because L9H9 and L9H6 attend to it with high attention, copying the indirect object.''

\textit{LLM:} ``GPT-2 predicts `Mary' because L9H9 attends to Mary with 66.5\% attention while giving John only 7.0\%, identifying Mary as the indirect object recipient.''

\subsection{RQ3: Failure Analysis}
\label{sec:results:rq3}

\begin{figure}[t]
    \centering
    \includegraphics[width=1.0\columnwidth]{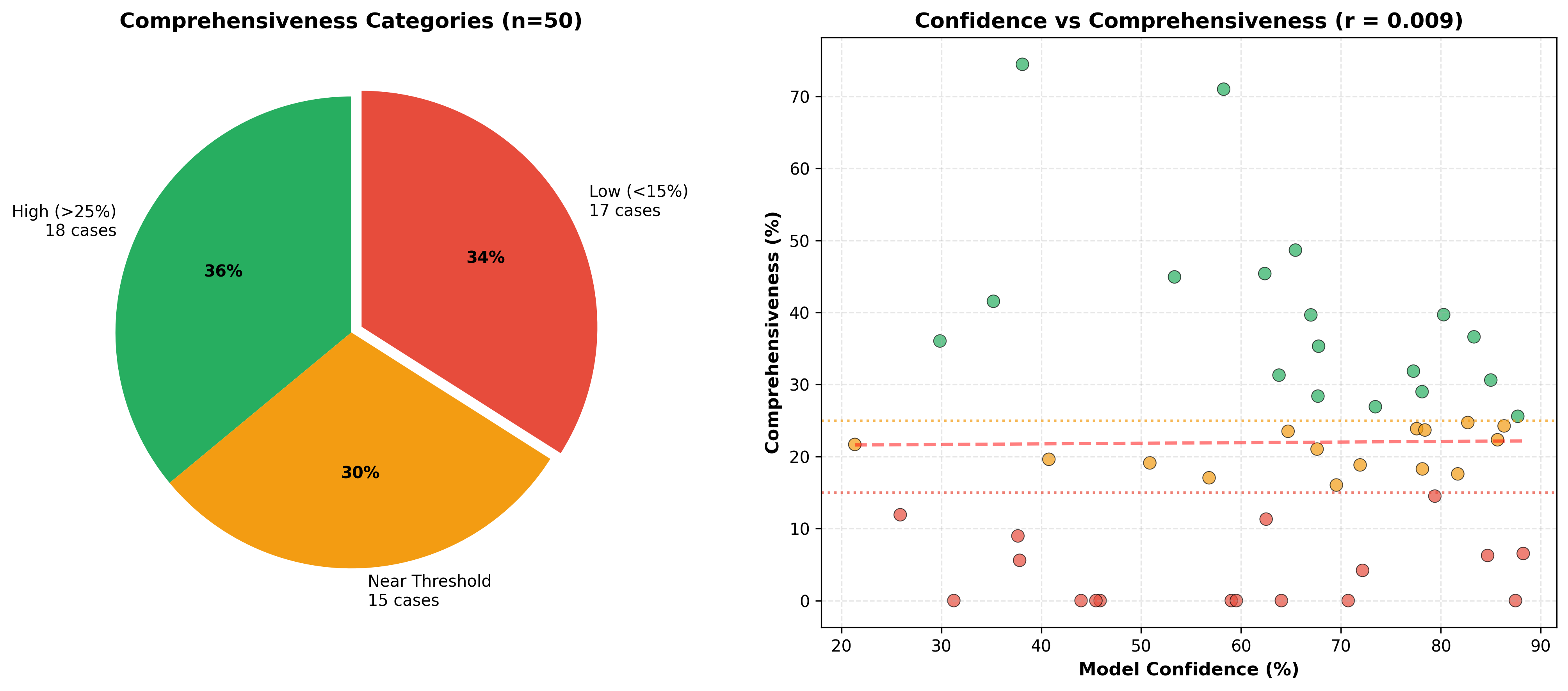}
     \caption{Comprehensiveness distribution across 50 prompts. 34\% of cases show low comprehensiveness ($<$15\%), indicating explanation-mechanism divergence.}
     \label{fig_06a_failure_taxonomy}
\end{figure}

Figure~\ref{fig_06a_failure_taxonomy} shows the comprehensiveness distribution: 34\% of cases exhibit low comprehensiveness ($<$15\%), 30\% near threshold (15--25\%), and 36\% high ($>$25\%).

Analysis of low-comprehensiveness cases reveals three failure categories:

\begin{enumerate}[leftmargin=1.5em, itemsep=1pt, topsep=2pt]
    \item \textbf{Distributed Computation.} Behavior emerges from many heads with moderate contributions. No small subset dominates.
    \item \textbf{Missing Cited Head.} Top contributors for specific prompts are not in the fixed circuit. L10H10 appears in 82\% of failure cases but is absent from our top-6.
    \item \textbf{Redundant Head Activity.} Heads are active but do not increase causal coverage when added.
\end{enumerate}

\textbf{Confidence does not predict faithfulness.}\quad Correlation between model confidence and comprehensiveness is $r = 0.009$. High-confidence predictions can rely on distributed mechanisms poorly captured by explanations.

\textbf{Adding heads does not help.}\quad Adding L10H10 to the circuit drops F1 from 36.0\% to 34.4\%, indicating redundancy rather than missing causal structure.


\section{Discussion}
\label{sec:discussion}

\textbf{The sufficiency-comprehensiveness gap.}\quad 100\% sufficiency with 22\% comprehensiveness reveals that cited heads are sufficient but not necessary. The model maintains backup mechanisms, suggesting transformers implement redundant computation that makes them robust to ablation but difficult to explain concisely.

\textbf{LLM-generated explanations.}\quad The 66\% quality improvement demonstrates that LLMs can translate circuit data into readable descriptions while maintaining mechanistic grounding. This scales better than templates as circuit complexity grows.

\textbf{Implications for trust.}\quad The lack of correlation between confidence and comprehensiveness ($r = 0.009$) means users cannot infer explanation quality from model confidence. Systems should report comprehensiveness alongside explanations.

\textbf{Limitations.}\quad (1)~Single task (IOI only). (2)~Single model (GPT-2 Small, 124M). (3)~No human evaluation of explanation utility. (4)~Fixed circuit (top-6 heads; adaptive selection might help). (5)~LLM dependency adds cost. (6)~Global head importance (dataset average) vs.\ instance-specific. (7)~No gradient baselines (Integrated Gradients, SHAP). (8)~50 prompts; larger evaluation would increase confidence.

\textbf{Future work.}\quad Multi-task evaluation (Greater-Than, induction heads), larger models (GPT-2 Medium/Large), adaptive per-instance circuits, human studies, and distilled explanation models. Additionally, we aim to integrate distributional faithfulness metrics, such as those proposed by \citet{yeo2025faithful} in the future.


\section{Conclusion}
\label{sec:conclusion}

We presented a pipeline for generating causally faithful NL explanations from mechanistic circuit analysis. On IOI in GPT-2 Small, we identified six attention heads accounting for 61.4\% of the logit difference. Adapting ERASER metrics to circuit-level components, we found 100\% sufficiency but only 22\% comprehensiveness, revealing distributed backup mechanisms. LLM-generated explanations outperformed templates by 66\%. Failure analysis found no correlation between confidence and faithfulness ($r = 0.009$) and identified three divergence categories.

These findings show that faithful explanations require causal grounding beyond attention patterns, and that comprehensiveness gaps reveal fundamental properties of neural computation.


\section*{Impact Statement}

This paper presents work whose goal is to advance the field of Machine Learning, specifically the interpretability and trustworthiness of language models. Our pipeline for generating faithful explanations from mechanistic analysis could improve transparency in deployed AI systems. We specifically note that low comprehensiveness (22\%) means explanations capture only partial mechanisms. Deploying such explanations without this caveat could mislead users into overconfidence.

\section*{Acknowledgements}
I thank Prof. Dr. Ernst Georg Haffner for supervision and feedback on this work.


\bibliography{references}
\bibliographystyle{icml2026}


\newpage
\appendix
\onecolumn

\section{IOI Prompt Templates}
\label{app:prompts}

\begin{verbatim}
Template 1: "When {IO} and {S} went to 
             the store, {S} gave a drink to"
Template 2: "When {IO} and {S} went to 
             the park, {S} handed a flower to"
\end{verbatim}

Name pairs include: (Mary, John), (Alice, Bob), (Sarah, Tom), (Emma, James), (Lisa, David), (Anna, Michael), (Sophie, Daniel), (Rachel, Chris), (Laura, Kevin), (Julia, Peter), and 15 others.

\section{Complete Results}
\label{app:results}

\begin{table}[h]
\caption{Complete experimental results.}
\label{tab:full_results}
\vskip 0.1in
\centering
\begin{tabular}{lr}
\toprule
\textbf{Metric} & \textbf{Value} \\
\midrule
Model Accuracy              & 100\% \\
Clean Logit Difference      & 3.36 \\
Corrupt Logit Difference    & $-3.53$ \\
Circuit Coverage            & 61.4\% \\
Sufficiency (mean $\pm$ std) & 100.0\% $\pm$ 0.0\% \\
Comprehensiveness (mean $\pm$ std) & 22.0\% $\pm$ 17.3\% \\
F1 Score                    & 36.0\% \\
Confidence--Comp.\ Correlation & 0.009 \\
Template Quality            & 60\% \\
Learned Quality             & 99\% \\
Quality Improvement         & +66\% \\
Low Comp.\ Cases ($<$15\%) & 34\% \\
\bottomrule
\end{tabular}
\end{table}

\section{Additional Figures}
\label{app:figures}

\begin{figure}[h]
    \centering
    \includegraphics[width=1.0\textwidth]{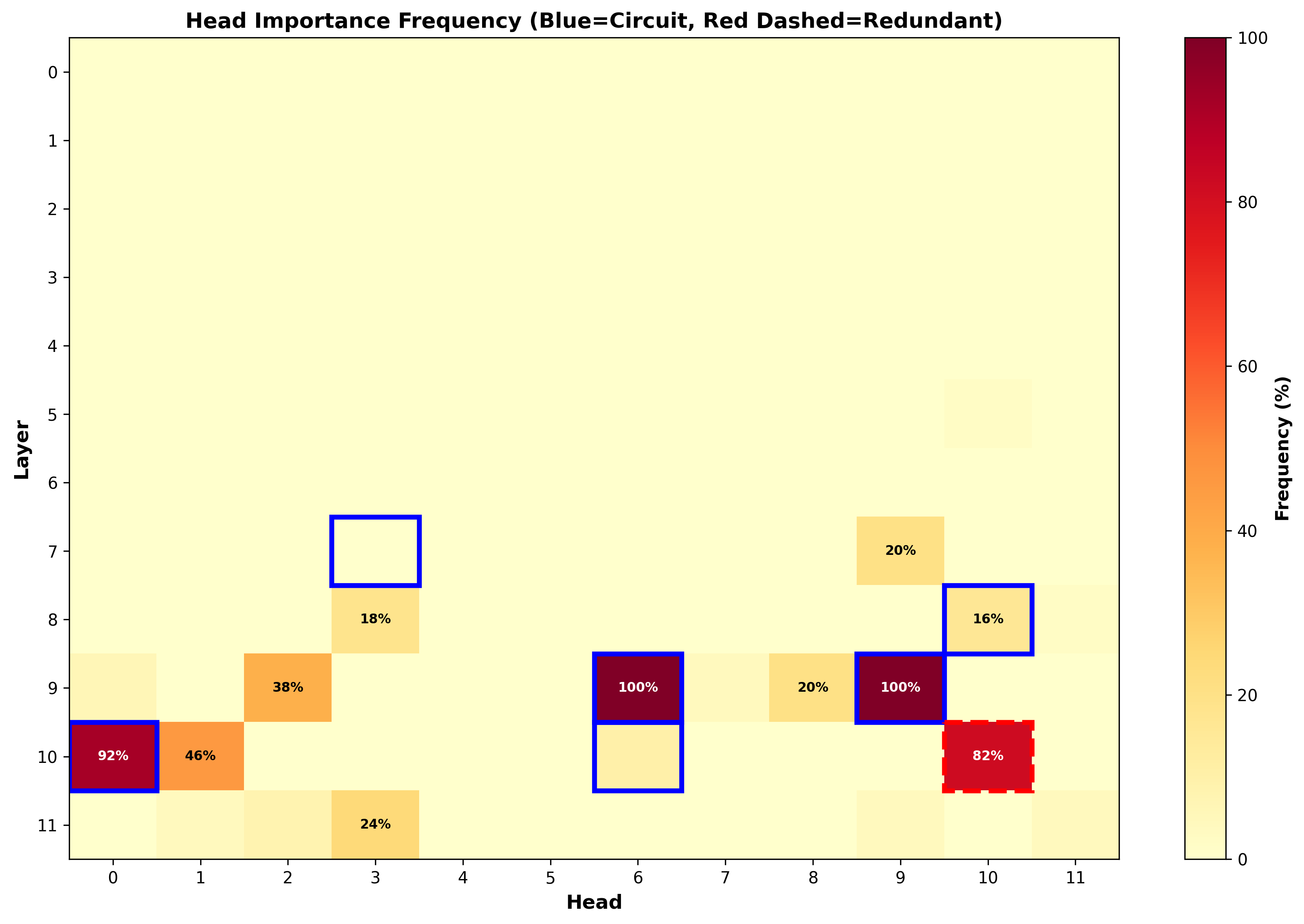}
    \caption{Detailed head contribution heatmap showing per-prompt importance of all 144 attention heads.}
    \label{fig:fig_06b_head_heatmap}
\end{figure}

\begin{figure}[h]
    \centering
    \includegraphics[width=1.0\textwidth]{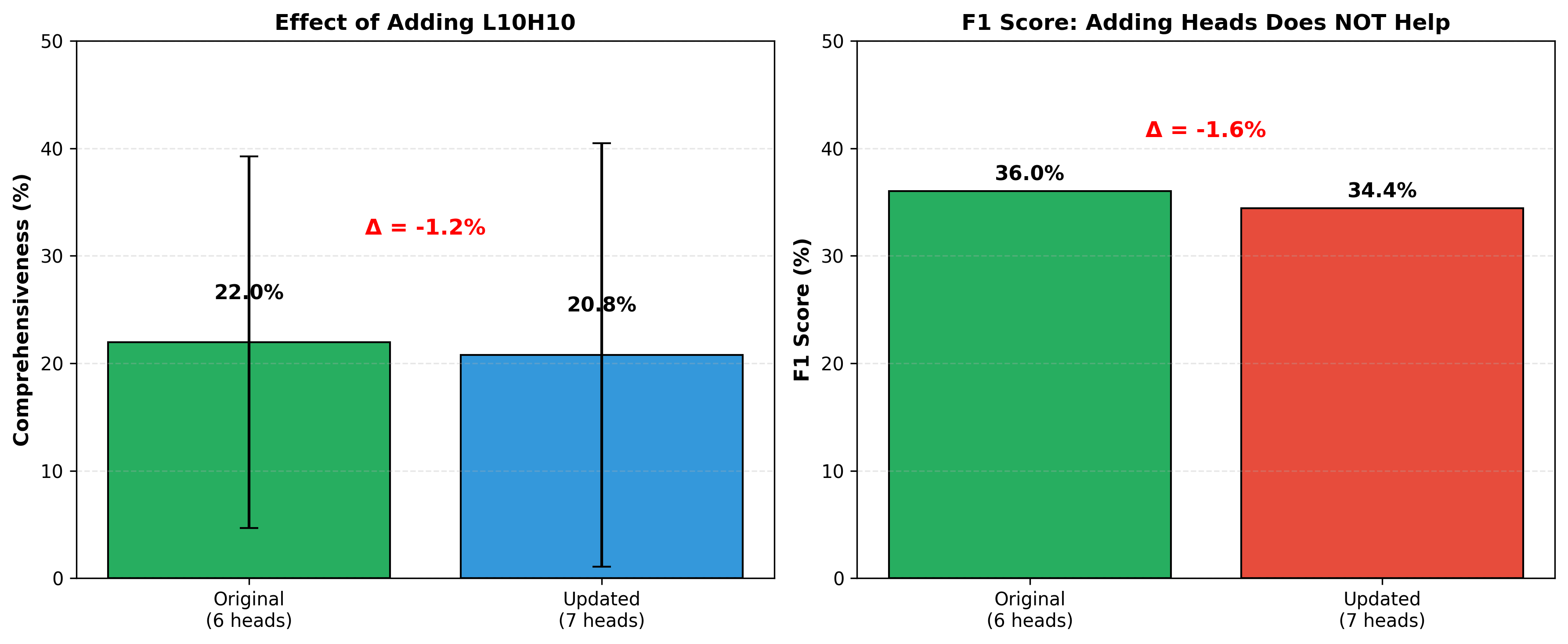}
    \caption{Effect of adding L10H10 to the circuit. F1 drops from 36.0\% to 34.4\%, confirming redundancy.}
    \label{fig:fig_06c_circuit_comparison}
\end{figure}

\section{Explanation Quality Criteria}
\label{app:quality}

Following e-SNLI patterns \citep{camburu2018esnli}, we score explanations on five criteria (each 0--1):

\begin{enumerate}
    \item \textbf{Mentions specific heads} (L9H9, L9H6, etc.)
    \item \textbf{Includes actual attention percentages} (e.g., 66.5\%)
    \item \textbf{Mentions the prediction correctly} (e.g., ``Mary'')
    \item \textbf{Mentions indirect object name} (e.g., ``Mary'')
    \item \textbf{Mentions subject name} (e.g., ``John'')
\end{enumerate}

Quality score = average across criteria.Template explanations score 0\% on criterion 2 (no actual percentages) and 0\% on criterion 5 (no subject name), yielding 3/5 = 60\%.
\end{document}